\documentclass[fleqn,10pt]{wlscirep}
\usepackage[utf8]{inputenc}
\usepackage[T1]{fontenc}
\usepackage{amsmath}    
\usepackage{amsfonts}   
\usepackage{amssymb}    
\usepackage{url}        
\usepackage{textcomp}   
\usepackage{placeins}
\usepackage{float}
\usepackage[inkscapelatex=false]{svg}

\title{Graph Transformer-based Pathway Embedding for Cancer Prognosis}

\author[1]{Koushik Howlader}
\author[2,*]{Md Tauhidul Islam}
\author[1,*]{Wei Le}
\affil[1]{Iowa State University, Iowa, USA}
\affil[2]{Stanford University, CA, USA}

\affil[*]{corresponding authors: weile@iastate.edu \& tauhid@stanford.edu}


\begin{abstract}
    Accurate prediction of cancer progression remains a challenge due to the high heterogeneity of molecular omics data across patients. While biologically informed models have improved the interpretability of these predictions, a persistent limitation lies in how they encode individual genes to construct pathway representations. Existing hierarchical models typically derive gene features by directly mapping raw molecular inputs, whereas integration frameworks often rely on simple statistical aggregations of patient-level signals. These approaches often fail to explicitly learn a shared base representation for each gene, thereby limiting the expressiveness and biological accuracy of downstream pathway embeddings. To address this, we introduce PATH, a modulation-based, patient-conditioned gene embedding strategy. PATH represents a paradigm shift by starting from a shared base embedding for each gene, preserving a stable biological identity across the population, and then dynamically adapting it using patient-specific copy number variation (CNV) and mutation signals. This allows the model to capture subtle individual molecular variations while maintaining a consistent latent understanding of the gene itself. We integrate PATH into a graph transformer framework that models interactions among biologically connected pathways through pathway-guided attention. Across pan-cancer metastasis prediction, PATH achieves an F1 score of 0.8766, representing an 8.8\% improvement over the current state-of-the-art multi-omics benchmarks. Beyond superior predictive accuracy, our approach identifies biologically meaningful pathways and, crucially, reveals disease-state-specific pathway rewiring, offering new insights into the evolving pathway–pathway interactions that drive cancer progression.

\end{abstract}

\begin{document}

\flushbottom
\maketitle

%
%


\section{Introduction}

Recent advances in multi-omics technologies have enabled the study of cancer progression in much greater detail \cite{hasin2017}. While these data provide a more comprehensive view of gene regulation, their high dimensionality and strong heterogeneity across patients create major challenges for accurately predicting clinical outcomes such as cancer progression and metastasis \cite{mcgranahan2017,tarazona2021}. To address this, a wide range of computational methods, including machine learning and deep learning, have been developed to analyze genomic and multi-omics data and build predictive models \cite{islam2020integrative,fu2020swine,wang2021mogonet,tanvir2024mogat}. However, many of these approaches still struggle to capture the most informative features from heterogeneous cancer data while preserving biologically meaningful structure \cite{tarazona2021,liu2024pathformer}.

To bridge the gap between predictive accuracy and biological understanding, recent research has increasingly shifted toward biologically informed deep learning models \cite{elmarakeby2021pnet,oh2021pathcnn,ma2024graphpath,liu2024pathformer}. These frameworks incorporate prior biological knowledge, such as gene-pathway memberships and pathway-pathway interactions, directly into their architectures \cite{elmarakeby2021pnet,wang2021mogonet,tanvir2024mogat,ma2024graphpath,liu2024pathformer}. By treating biological pathways as functional units rather than independent collections of genes, such models have demonstrated improved interpretability and strong predictive performance in disease classification, prognosis, and molecular stratification \cite{elmarakeby2021pnet,oh2021pathcnn,ma2024graphpath,liu2024pathformer}.

However, a persistent limitation of current state-of-the-art models lies in how they encode gene-level information to construct pathway representations. Existing hierarchical models often rely on biologically constrained sparse connections in which gene nodes are driven directly by patient molecular measurements such as mutation or copy-number profiles \cite{elmarakeby2021pnet}. Likewise, several graph-based multi-omics integration frameworks operate on high-dimensional omics feature vectors or patient-similarity graphs, which can be effective for prediction but do not explicitly learn a stable latent biological identity for each gene across patients \cite{wang2021mogonet,tanvir2024mogat}. Furthermore, recent transformer-based approaches often compress multimodal inputs into gene-level representations using summary statistics before pathway aggregation, which may obscure subtle biological variation important for distinguishing aggressive disease states \cite{liu2024pathformer}. More broadly, although pathway-aware models have advanced biological interpretability, the problem of learning patient-conditioned gene representations before pathway aggregation remains insufficiently explored \cite{oh2021pathcnn,ma2024graphpath}.

In this work, we introduce PATH, a novel modulation-based, patient-conditioned gene embedding strategy that preserves stable gene identity while capturing individual molecular variation. PATH represents a significant methodological shift 
\FloatBarrier
\begin{figure*}[t!]
\centering
\includegraphics[width=\linewidth]{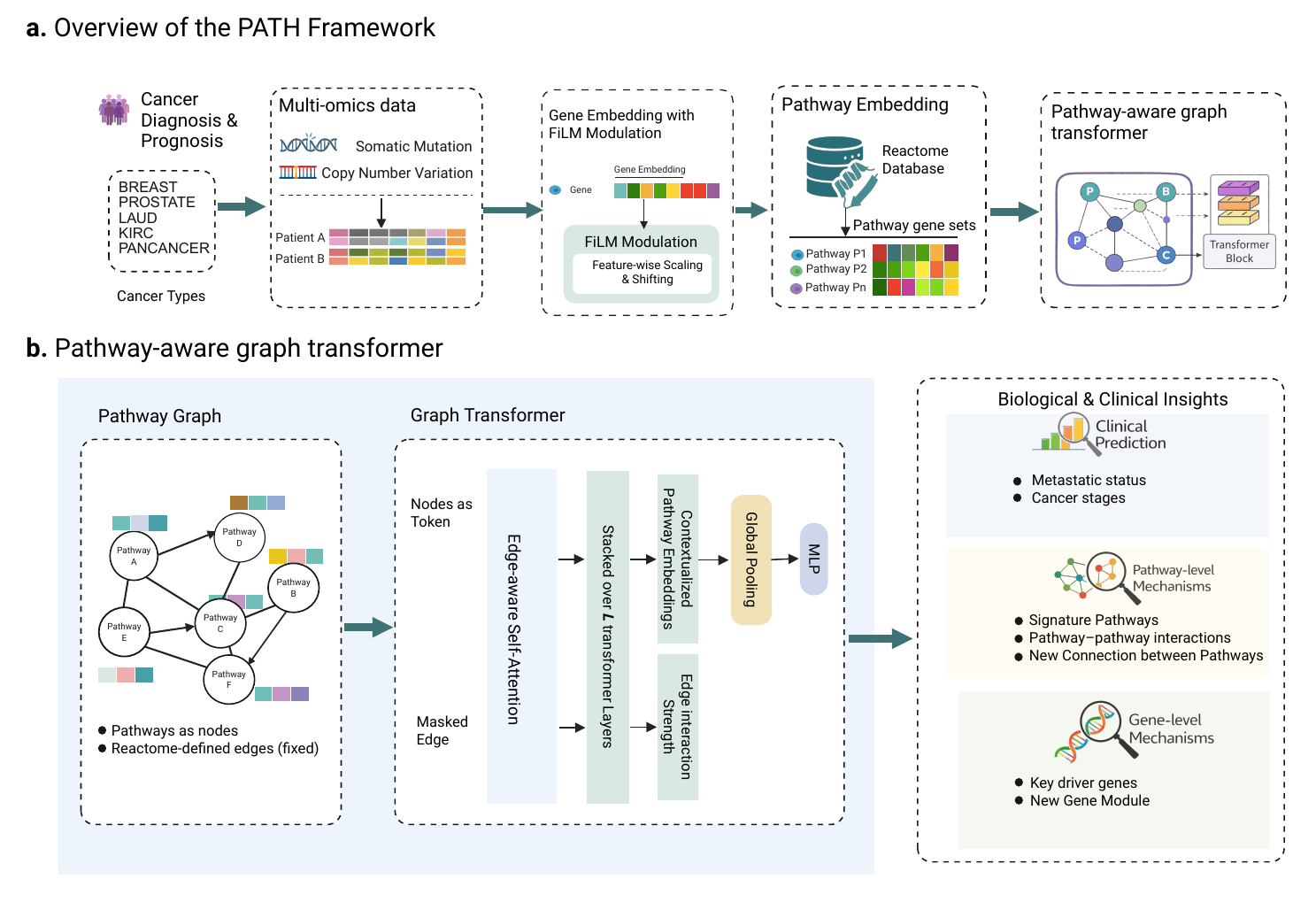}
\caption{\textbf{Overview of the PATH framework and model core.}
\textbf{(a)} For each cancer cohort, patient multi-omics profiles (somatic mutation and copy-number variation) are used as input. PATH performs \emph{gene-level representation learning} to produce patient-specific gene features, aggregates these features into \emph{pathway embeddings} using Reactome gene sets, and generates pathway tokens summarizing pathway activity. The pathway tokens are passed to a \emph{pathway-aware graph transformer} to model pathway crosstalk and produce predictions and interpretable outputs.
\textbf{(b)} In the model core, pathways are treated as nodes in a pathway graph, with Reactome-defined edges providing a biological prior. The transformer performs contextualized, edge-aware self-attention over pathway tokens; soft masked edges improve robustness to incomplete pathway knowledge while learning data-driven interaction strengths. Stacking $L$ layers yields contextualized pathway embeddings, which are pooled and passed to an MLP for clinical prediction. Learned attention and interaction strengths enable interpretability, including signature pathways, pathway--pathway interactions, model-inferred connections, and key driver genes.}
\label{fig:overview}
\end{figure*}

\FloatBarrier
by starting from a shared base embedding for each gene, ensuring that the model maintains a universal latent representation of the gene as a biological entity. This base representation is then dynamically adapted through a modulation process using patient-specific signals, such as copy number variations (CNV) and mutation counts. These expressive, modulated gene embeddings are aggregated into pathway embeddings, providing a more robust foundation for downstream learning tasks.

We integrate PATH into a graph transformer framework to model the complex interactions among biologically connected pathways. By leveraging pathway-guided attention derived from a curated pathway knowledgebase, the model moves beyond static hierarchies to capture the dynamic crosstalk among biological processes \cite{gillespie2022reactome,prahallad2016}. This allows the model to capture how pathways interact across different disease states. To provide additional flexibility, we introduce a soft-mask mechanism that relaxes strict pathway connectivity, enabling the model to refine known interactions and explore potential new pathway--pathway relationships.

We evaluate the proposed framework on cancer progression prediction tasks, including metastasis and stage (early vs.\ late) prediction. In addition to predictive performance, the framework provides biologically meaningful insights into pathway activity and interaction patterns associated with disease progression. In particular, it reveals pathways and gene sets linked to cancer progression, disease-state-specific pathway rewiring, and potential new pathway--pathway interactions. These results demonstrate that learning patient-specific gene representations can improve both prediction accuracy and biological interpretability. In the following sections, we present a detailed evaluation of the model and analyze the insights it provides into cancer progression.
\section{Results}

\subsection{Overview of PATH and outputs}
PATH is designed to improve pathway-based genomic prediction while providing built-in interpretability at three levels: genes, pathways and pathway–pathway interactions. Given patient multi-omics inputs, including somatic mutations and copy number variation, the model first learns gene-level representations, then aggregates genes into pathway tokens, and finally models pathway crosstalk using a pathway-aware graph transformer (Fig.~\ref{fig:overview}a). Pathway definitions were curated from the Reactome database. In this framework, pathways are treated as nodes and known interactions between pathways as edges, forming a pathway graph representation. Within the model, pathways are represented as tokens, and edge-aware self-attention over masked edges enables PATH to refine these prior relationships and learn context-dependent interaction strengths from the data (Fig.~\ref{fig:overview}b). As a result, PATH produces standard clinical predictions, such as metastatic status or cancer stage, together with mechanistic readouts, including ranked pathways, candidate driver genes, inferred pathway–pathway interaction patterns and model-inferred new pathway connections (Fig.~\ref{fig:overview}b).

This multiscale design structures the Results section. We begin with the pan-cancer primary-versus-metastatic task, which represents the most heterogeneous setting because it integrates tumours from multiple cancer types with distinct tissue origins, molecular landscapes and progression trajectories. We then evaluate PATH on a disease-focused prostate cancer cohort using the same primary-versus-metastatic classification task, followed by BLCA early-versus-late-stage classification. For each dataset, we first compare predictive performance against established baseline models and then examine the biological insights recovered by PATH at both the pathway and gene levels.

\subsection{PATH captures metastatic progression across heterogeneous pancancer samples}
\begin{figure*}[t!]
  \centering
  \includegraphics[width=\linewidth]{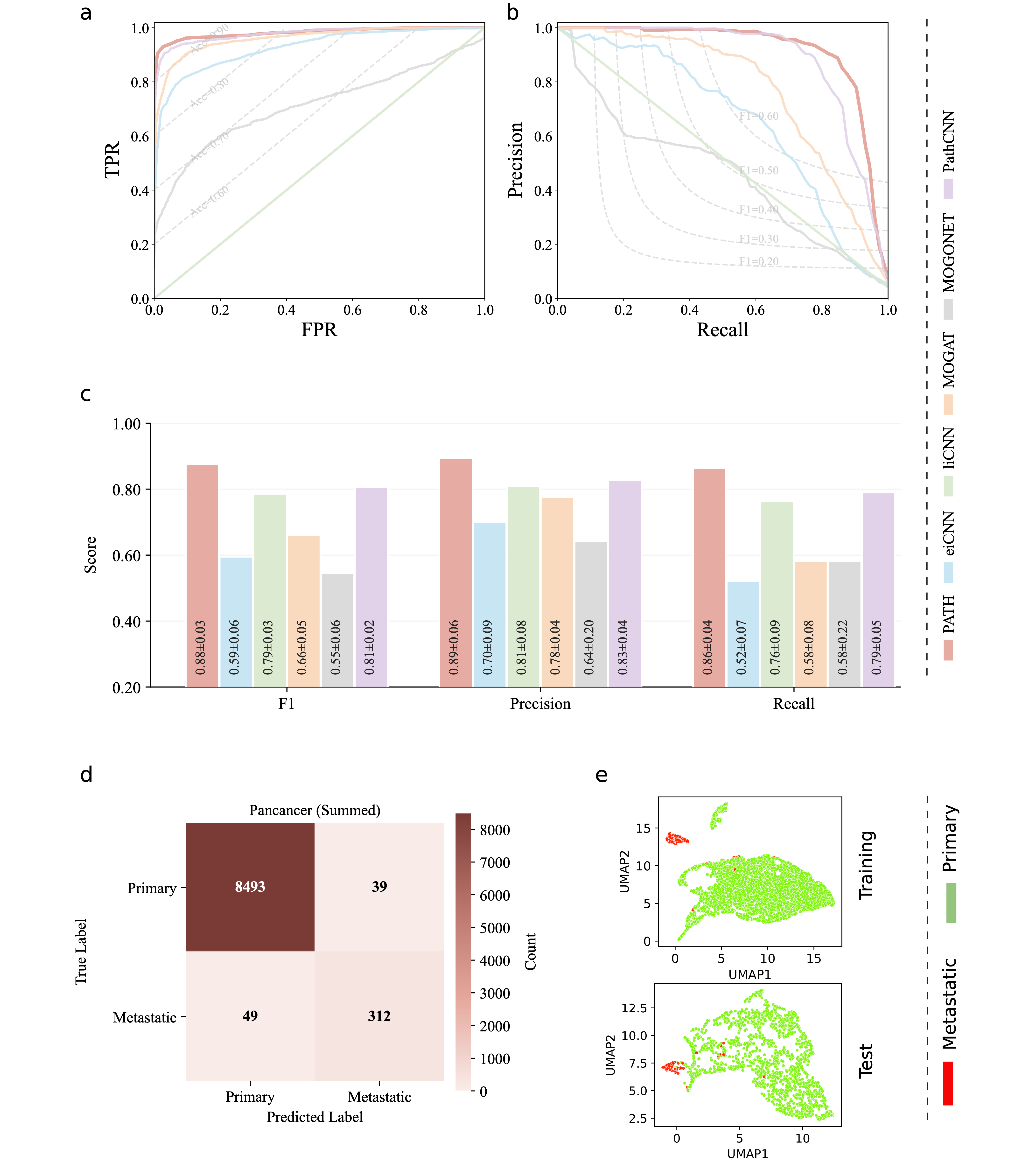}
  \caption{\textbf{PATH accurately predicts metastatic status in the pancancer cohort and learns progression-associated latent representations.}
  \textbf{(a)} ROC curves comparing PATH with baseline models for primary versus metastatic classification in the pancancer dataset.
  \textbf{(b)} PR curves showing that PATH preserves favorable precision across a broad recall range under class imbalance.
  \textbf{(c)} Mean $\pm$ standard deviation of F1 score, precision, and recall across five-fold cross-validation, demonstrating superior overall performance of PATH relative to competing methods.
  \textbf{(d)} Confusion matrix summarizing PATH predictions across pancancer samples.
  \textbf{(e)} UMAP projections of learned latent representations for the training and test sets, with samples colored by progression state, showing separation between primary and metastatic tumors.
  These results indicate that PATH provides robust pancancer metastatic prediction and learns biologically meaningful representations associated with tumor progression.}
  \label{fig:pancancer_performance}
\end{figure*}

We first analyzed the pan-cancer cohort, the most heterogeneous setting in this study, because it combines multiple cancer types into a single primary-versus-metastatic classification task. As shown in Fig.~\ref{fig:pancancer_performance}a, PATH achieved the strongest overall discrimination in the ROC analysis and maintained high precision across a broad recall range in the precision–recall analysis (Fig.~\ref{fig:pancancer_performance}b). Across five-fold cross-validation, PATH achieved the highest F1 score ($0.88 \pm 0.03$), precision ($0.89 \pm 0.06$), and recall ($0.86 \pm 0.04$), exceeding all competing pathway-based and multi-omics baselines (Fig.~\ref{fig:pancancer_performance}c). In addition, the confusion matrix in Fig.~\ref{fig:pancancer_performance}d further confirmed balanced performance across both classes, with 8,493 primary and 312 metastatic samples correctly classified, compared with only 39 primary and 49 metastatic misclassifications. Finally, the UMAP visualization in Fig.~\ref{fig:pancancer_performance}e showed clear class-wise separation in the learned embeddings for both the training and test sets, providing a qualitative complement to the held-out predictive performance. See complete UMAP visualizations for the pancancer and prostate cohorts in Supplementary Figs.~S1 and S2, respectively.

\FloatBarrier
\FloatBarrier

\FloatBarrier
\begin{figure*}[t!]
  \centering
  \includegraphics[width=\linewidth]{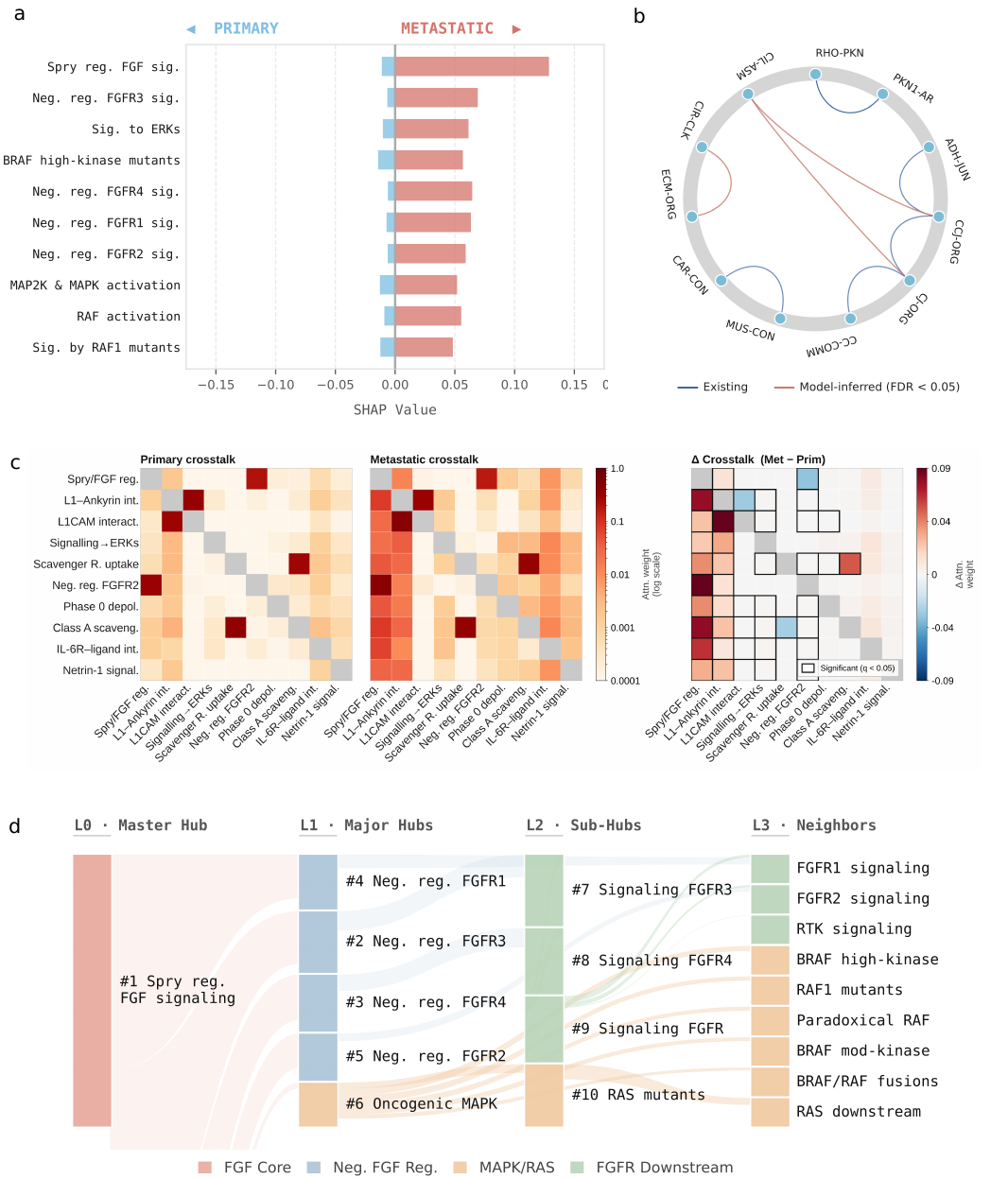}
  \caption{
  \textbf{(a)} SHAP-based pathway analysis highlighting the top pathways contributing to primary versus metastatic classification, with mean absolute SHAP values summarizing overall pathway importance.
  \textbf{(b)} Circular network of model-inferred novel pathway--pathway interactions (FDR $<$ 0.05) absent from the reference Reactome network, shown alongside existing curated connections.
  \textbf{(c)} Pathway--pathway attention heatmaps for primary and metastatic tumors, together with the differential crosstalk matrix ($\Delta$ = Met $-$ Prim), showing rewiring of pathway co-activation during metastatic progression.
  \textbf{(d)} Hierarchical hub analysis of the learned pathway network, identifying an FGF/FGFR- and MAPK-centered signaling architecture with \textit{Sprouty regulation of FGF signaling} as the master hub across four levels (L0--L3).
  Together, these results show that PATH captures metastatic progression through coordinated pathway importance patterns, novel inter-pathway relationships, and network-level reorganization of signaling programs.}
  \label{fig:pancancer_pathway_analysis}
\end{figure*}
\FloatBarrier

\FloatBarrier
\begin{table*}[t]
\centering
\caption{\textbf{Supplementary Table S1. Top pathways identified by PATH for metastatic progression (Figure 3a).} 
Pathways are ranked based on SHAP importance and validated using literature evidence.}
\label{tab:fig3a_pathways}
\small
\begin{tabular}{p{0.8cm} p{4.5cm} p{2.0cm} p{1.5cm} p{7.0cm}}
\hline
\textbf{Rank} & \textbf{Pathway} & \textbf{Evidence Type} & \textbf{Reference} & \textbf{Comment} \\
\hline

1 & Sprouty regulation of FGF sig. 
& Indirect 
& \cite{Kawazoe2019Sprouty} 
& Loss of Sprouty feedback prolongs FGFR--MAPK signaling, increasing tumor cell growth and migration that can lead to metastasis. \\

2 & Neg. regulation of FGFR3 sig. 
& Indirect 
& \cite{diMartino2012FGFR3} 
& Reduced control of FGFR3 signaling enhances survival and proliferation signals that may support tumor spread. \\

3 & Sig. to ERKs 
& Direct 
& \cite{Kciuk2022MAPK} 
& ERK activation drives proliferation, survival, and migration, directly enabling cancer cells to invade and metastasize. \\

4 & BRAF high kinase activity mutants 
& Direct 
& \cite{Bahar2023MAPK} 
& Hyperactive BRAF strongly increases MAPK signaling, promoting aggressive tumor behavior and metastatic progression. \\

5 & Neg. regulation of FGFR4 sig. 
& Indirect 
& \cite{Levine2020FGFR4, GarciaRecio2020FGFR4} 
& Loss of FGFR4 regulation enhances signaling linked to invasion and aggressive tumor behavior. \\

6 & Neg. regulation of FGFR1 sig. 
& Indirect 
& \cite{Tomlinson2012FGFR1, Fan2024FGFR1} 
& Reduced inhibition of FGFR1 promotes EMT and invasion, enabling cancer cells to spread. \\

7 & Neg. regulation of FGFR2 sig. 
& Indirect 
& \cite{Fan2024FGFR1} 
& Disruption of FGFR2 regulation may alter growth signaling balance, indirectly supporting tumor progression. \\

8 & MAP2K and MAPK activation 
& Direct 
& \cite{Bahar2023MAPK, Kciuk2022MAPK} 
& Activates core signaling cascade that promotes tumor growth, invasion, and metastatic transition. \\

9 & RAF activation 
& Direct 
& \cite{Bahar2023MAPK} 
& RAF activation amplifies MAPK signaling, increasing tumor cell proliferation and metastatic potential. \\

10 & Sig. by RAF1 mutants 
& Indirect 
& \cite{Bahar2023MAPK} 
& Aberrant RAF1 signaling enhances tumor growth and may contribute to metastasis in specific contexts. \\

\hline
\end{tabular}
\end{table*}
\FloatBarrier
\FloatBarrier

\begin{table*}[t]
\centering
\caption{\textbf{Supplementary Table S2. Top genes identified by PATH for metastatic progression (Figure 4c).}
Genes are ranked based on SHAP importance and validated using literature evidence.}
\label{tab:fig3b_genes}
\small
\begin{tabular}{p{0.8cm} p{1.5cm} p{1.5cm} p{1.5cm} p{10.0cm}}
\hline
\textbf{Rank} & \textbf{Gene} & \textbf{Evidence Type} & \textbf{Reference} & \textbf{Comment} \\
\hline
1 & AR
& Direct
& \cite{quigley2018ar, abida2019genomic}
& AR is amplified or mutated in the majority of metastatic tumors, allowing it to remain active even after androgen deprivation and drive continued tumor growth and spread. \\
2 & RPTOR
& Indirect
& \cite{hsieh2012mtor, shorning2020pi3k}
& As the core scaffold of mTORC1, RPTOR amplification boosts protein synthesis programs that enable cancer cells to invade surrounding tissue and metastasize. \\
3 & POLR3A
& Indirect
& \cite{colis2019polr3a}
& POLR3A keeps aggressive prostate cancer cells in a proliferative, undifferentiated state by sustaining the high tRNA output needed for rapid tumor growth. \\
4 & EIF4E
& Direct
& \cite{furic2010eif4e, hsieh2012mtor}
& EIF4E drives the translation of proteins that help cancer cells invade and spread; its activity rises as tumors progress to castration-resistant disease. \\
5 & AKT1
& Direct
& \cite{gonzalez2023pten, lorente2025clinical}
& AKT1 promotes tumor cell survival and motility downstream of PI3K; its amplification or mutation is linked to faster progression and worse survival in metastatic disease. \\
6 & \textcolor{red}{PPP1R12B}
& Hypothesis
& \cite{wo2024mechanism}
& PPP1R12B Identified as a fibroblast-associated gene linked to prostate cancer progression and immune interactions, but its direct role in metastatic progression remains unvalidated and requires further investigation. \\
7 & NFKBIA
& Direct
& \cite{mccall2012nfkb, nadiminty2015nfkb}
& Loss of NFKBIA removes the brake on NF-$\kappa$B, unleashing inflammatory signaling that promotes bone metastasis and resistance to androgen-targeting therapy. \\
8 & POLR3G
& Indirect
& \cite{colis2019polr3a, arimbasseri2022polr3g}
& POLR3G re-emerges in cancer cells under MYC activation and keeps them in a stem-like, proliferative state that supports tumor growth and spread. \\
9 & PTEN
& Direct
& \cite{taylor2010prad, gonzalez2023pten}
& PTEN loss disables the main brake on PI3K--AKT signaling, accelerating tumor growth, invasion, and progression to castration-resistant metastatic disease. \\
10 & PTPN11
& Direct
& \cite{zhang2016shp2}
& PTPN11 (SHP2) disrupts normal cell polarity and triggers epithelial-to-mesenchymal transition in prostate cancer cells, making them more motile and prone to metastasize. \\
\hline
\end{tabular}
\end{table*}
\FloatBarrier

\subsection{PATH-derived pathway analysis of pancancer metastatic progression}
In the previous section, we demonstrated the predictive performance of PATH on the pan-cancer dataset. We next examined the biological insights revealed by the model. We hypothesized that alterations in pathway–pathway interactions contribute to cancer progression. As a first step, we show that PATH identifies the pathways most strongly associated with disease progression (Figure~\ref{fig:pancancer_pathway_analysis}\textbf{a}). We then analyse how pathway–pathway interaction patterns are rewired during cancer progression and whether these changes may explain the transition between disease states (Figure~\ref{fig:pancancer_pathway_analysis}\textbf{c}). In addition, we show that PATH identifies new pathway–pathway connections (Figure~\ref{fig:pancancer_pathway_analysis}\textbf{b}) and pathway hubs (Figure~\ref{fig:pancancer_pathway_analysis}\textbf{d}). Finally, we assess the biological plausibility of these findings using evidence from the literature shown in table S*.

Figure~\ref{fig:pancancer_pathway_analysis}\textbf{a} shows that \textbf{PATH identifies dominant metastatic signaling pathways}. To interpret these pathways, we ranked pathway-level SHAP attributions according to the mean absolute SHAP difference between the primary and metastatic classes. Owing to space limitations, pathway names are shown in abbreviated form in the main figure; the full pathway names are provided in Supplementary Fig.~S*. Here, we present the top 10 pathways, whereas the top 20 pathways are shown in Supplementary Fig.~S*. The complete pathway ranking is also provided in Supplementary File S* in CSV format. Among the top 10 pathways, eight have literature support for roles in metastatic progression. The pathways Neg. reg. FGFR4 sig. and Neg. reg. FGFR1 sig. do not directly indicate metastatic progression on their own. However, prior studies suggest that loss or dysregulation of negative feedback control around FGFR1/FGFR4 may contribute to metastatic progression by enabling stronger or more sustained FGFR signaling\cite{szybowska2021negative,fgfr1,tang2018fgfr4}.



In Figure~\ref{fig:pancancer_pathway_analysis}\textbf{a}, we showed the top-ranked pathways associated with metastatic progression. We next asked whether metastatic tumours exhibit rewiring of pathway–pathway crosstalk. Metastatic progression involves coordinated changes across interacting pathways rather than isolated alterations in single pathways. Interactions between pathways are present in both primary and metastatic tumours, but the pattern of pathway coordination changes during disease progression and may contribute to metastatic transition. Consistent with this idea, comparison of pathway–pathway crosstalk between primary and metastatic tumours revealed substantial rewiring of attention-based co-activation patterns (Fig.~\ref{fig:pancancer_pathway_analysis}\textbf{c}). The delta crosstalk map (Met $-$ Primary) identified both gained and lost pathway interactions in metastatic tumours. Cells outlined in black denote statistically significant edges ($q < 0.05$), confirming that these rewiring events are unlikely to be random.


\subsubsection*{Model-Inferred Novel Pathway Interactions}


Beyond known Reactome connections, PATH identified three novel pathway to pathway
interactions (FDR $<$ 0.05) absent from the reference Reactome network (Fig.~\ref{fig:pancancer_pathway_analysis}\textbf{b}) . These
edges were inferred by learning co activation weights between pathway pairs
from SHAP-derived pathway activity scores across primary and metastatic tumours,
retaining only those absent from the reference network with a base weight of zero. The first links \textbf{Extracellular Matrix Organization} to the 
\textbf{Circadian Clock}. ECM stiffness modulates circadian amplitude 
via integrin and Rho/ROCK signalling~\cite{streuli2019}, while the clock 
reciprocally controls rhythmic collagen secretion and MMP 
activity~\cite{meng2023}, a bidirectional axis relevant to tumour 
microenvironment remodelling. The remaining two edges connect both \textbf{Cell Junction Organization} 
and \textbf{Cell-cell Junction Organization} to \textbf{Cilium Assembly}. 
Cell junctions and primary cilia are co-regulated through shared polarity 
proteins and actin dynamics~\cite{tates2018,sfakianos2007}. Their concurrent 
dysregulation, in which junctional breakdown drives EMT~\cite{holst2011} and 
cilia loss is widespread across epithelial malignancies~\cite{wong2025}, 
supports these edges as coupled events in the metastatic programme.



\subsubsection*{Hierarchical hub analysis reveals an FGFR-centered metastatic signaling architecture}

Using \textbf{PATH}, we identified a compact signaling hierarchy centered on \textit{Spry regulation of FGF signaling}, with strong links to pathways involved in negative regulation of FGFR1, FGFR2, FGFR3 and FGFR4 signaling, and additional hubs in oncogenic MAPK, pan-FGFR and RAS-mutant signaling (Fig.~\ref{fig:pancancer_pathway_analysis}\textbf{d}). This organization is biologically plausible because Sprouty proteins are established modulators of receptor tyrosine kinase and MAPK/ERK signaling \cite{masoumimoghaddam2014sprouty}, dysregulated FGFR signaling is widely implicated in cancer progression and therapeutic resistance \cite{yue2021fgfr}, and sustained FGFR--MAPK signaling has been linked to invasive and metastatic behavior \cite{suyama2002ncadfgfr,xu2017spry2}. The prominence of MAPK- and RAS-related hubs is also consistent with the broader role of MAPK signaling in metastasis \cite{mapk}. At the same time, the specific hierarchy recovered by PATH, particularly the organization of FGFR subtype-specific regulatory pathways around a Spry-centered hub, is not directly established at pathway-pair resolution in the literature and therefore represents a biologically grounded, model-derived hypothesis for further experimental validation.

\FloatBarrier
\begin{figure*}[t!]
  \centering
  \includegraphics[width=\linewidth]{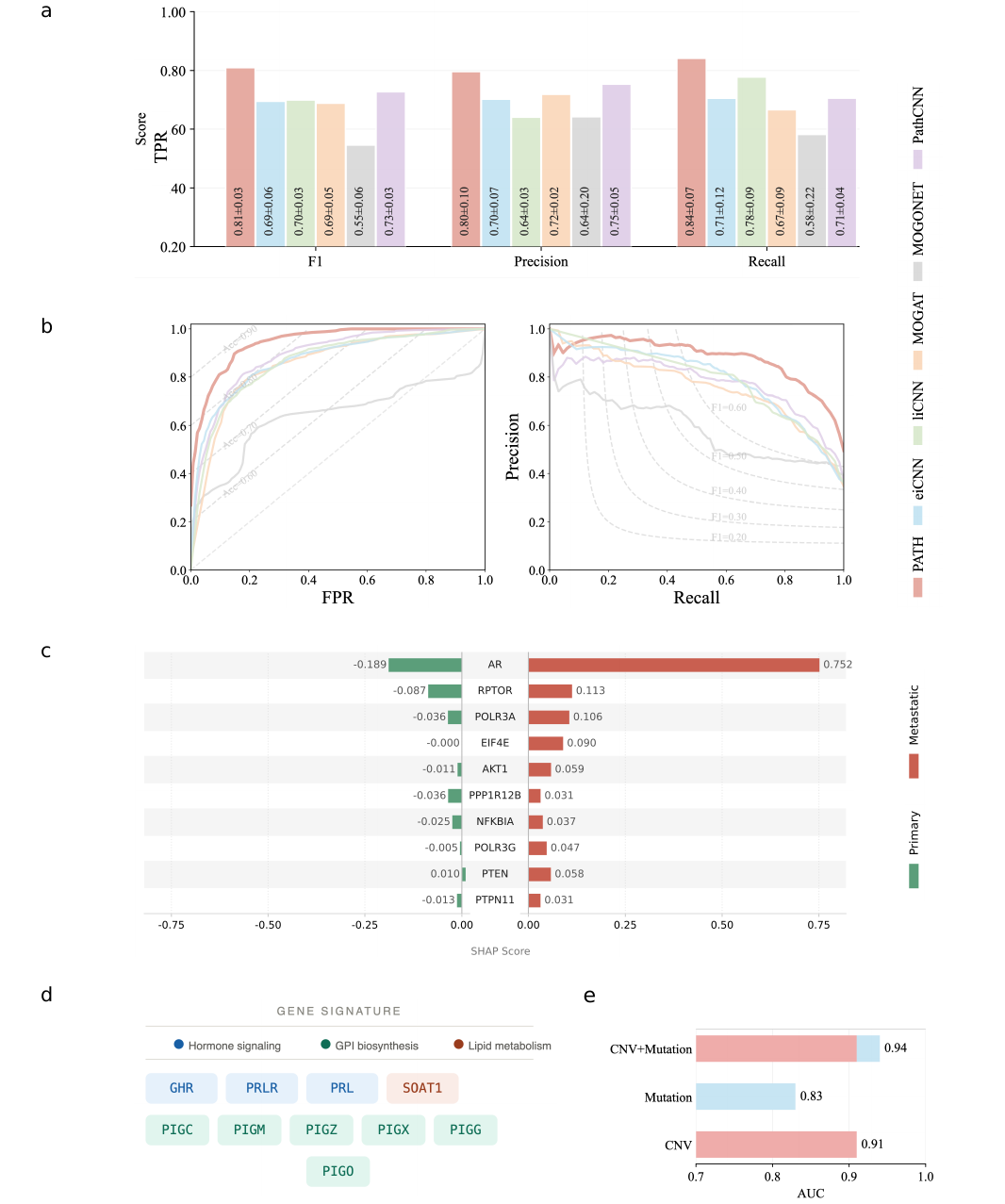}
  \caption{\textbf{PATH shows strong performance and interpretable biological signals in prostate cancer metastasis classification.}
  \textbf{(a)} F1 score, precision, and recall across models, showing the best overall performance by PATH.
  \textbf{(b)} ROC and precision--recall curves confirming the strong discriminative ability of PATH.
  \textbf{(c)} SHAP-based gene importance analysis identifying genes linked to primary and metastatic prostate cancer, with \textit{AR} contributing most strongly to the metastatic class.
  \textbf{(d)} Representative gene-signature modules highlighting hormone signaling, GPI biosynthesis, and lipid metabolism.
  \textbf{(e)} Ablation analysis showing that joint CNV and mutation integration gives the highest performance.
  Together, these results indicate that PATH combines improved prediction with biologically meaningful interpretation in prostate cancer metastasis.}
  \label{fig:prostate_cancer}
\end{figure*}
\FloatBarrier

\subsection{PATH outperforms baseline models in prostate cancer metastatic classification}
We next evaluated whether PATH transfers to a single cancer task with distinct genomic drivers by training on a prostate cancer
primary versus metastatic cohort. PATH outperformed all baselines across cross-validation metrics
(Fig.~\ref{fig:prostate_cancer}a,b), achieving the highest F1 score ($0.81\pm0.03$), precision ($0.80\pm0.10$), and recall
($0.84\pm0.07$). ROC and precision and recall curves confirmed superior discriminative ability
(Fig.~\ref{fig:prostate_cancer}b), indicating that the pathway-aware architecture transfers effectively from the pancancer setting
to a disease focused prediction task.

\subsubsection*{\textit{AR} emerges as the dominant gene-level driver of metastatic progression}
SHAP attribution analysis (Fig.~\ref{fig:prostate_cancer}c) identified the androgen receptor (\textit{AR}) as the single strongest metastatic
predictor (SHAP $=0.752$), more than six fold higher than the second ranked gene, \textit{RPTOR} ($0.113$). This dominance is expected given that AR pathway
reactivation is central to prostate cancer progression and mCRPC biology\textsuperscript{\cite{takeda2020,quigley2018}}. Because PATH SHAP scores derive from CNV
and mutation profiles, and AR copy number gain is among the most recurrent genomic events in metastatic cohorts, \textit{AR} is the expected dominant signal in this
data modality.

The remaining top ranked metastatic genes: \textit{RPTOR}, \textit{POLR3A}, \textit{EIF4E}, \textit{AKT1}, \textit{NFKBIA}, \textit{POLR3G}, \textit{PTEN}, and
\textit{PTPN11}, converge on mTOR signaling, translational control, and the PI3K/AKT/PTEN axis. \textit{RPTOR} (Raptor/mTORC1 scaffold) expression is elevated in
prostate tumors, and its knockdown attenuates migration and invasion in vitro\textsuperscript{\cite{xu2014}}. \textit{EIF4E}, the mTORC1-regulated cap-binding translation
initiator, is a convergence node for PI3K/Akt/mTOR and Ras/MAPK signaling; its phosphorylation correlates with prostate cancer disease progression in patient cohorts\textsuperscript{\cite{furic2010,dua2018}}.
\textit{AKT1} is the principal effector kinase downstream of PI3K, activated by PTEN loss\textsuperscript{\cite{cotter2018}}. \textit{PTEN} itself appears in both the metastatic and primary signatures:
homozygous deletion is disproportionately linked to metastatic and androgen independent progression, whereas hemizygous loss is the predominant primary tumor event\textsuperscript{\cite{yoshimoto2013}}.
\textit{PTPN11} encodes SHP2, which promotes prostate cancer metastasis by disrupting polarity complexes and driving EMT; elevated SHP2 expression correlates with metastasis and shortened survival\textsuperscript{\cite{zhang2015}}.

\subsubsection*{Gene module summaries yield interpretable progression associated programs}
To provide compact, biologically interpretable summaries of gene-level signals, we grouped key genes into three modules (Fig.~\ref{fig:prostate_cancer}d).

\paragraph{Hormone signaling module (\textit{GHR}, \textit{PRLR}, \textit{PRL}).}
\textit{GHR} mRNA is reported to be higher in prostate adenocarcinoma than in benign prostatic hyperplasia\textsuperscript{\cite{pandey2017}},
and GH/PRL axis activation can engage STAT5-related signaling in prostate cancer models\textsuperscript{\cite{chopin2004,goffin2010}}.
This module is therefore consistent with endocrine signaling crosstalk that can modulate tumor phenotypes.

\paragraph{GPI biosynthesis module (\textit{PIGC}, \textit{PIGM}, \textit{PIGZ}, \textit{PIGX}, \textit{PIGG}, \textit{PIGO}).}
These genes encode enzymes catalyzing successive steps in glycosylphosphatidylinositol (GPI) anchor preassembly. GPI-anchored proteins (GPI-APs) have established roles in cancer cell invasion and metastasis\textsuperscript{\cite{lauber2020}},
but the specific contribution of copy number alterations in these \textit{PIG} genes to prostate cancer metastasis has not been characterized; this module therefore represents a hypothesis generating observation requiring functional validation.

\paragraph{Lipid metabolism module (\textit{SOAT1}).}
SOAT1 (sterol O-acyltransferase 1) esterifies intracellular cholesterol for storage in lipid droplets. High \textit{SOAT1} expression in high-risk prostate cancer is associated with lymph node metastasis, elevated Gleason score,
and significantly shorter biochemical recurrence-free survival (HR $= 2.40$, $p < 0.001$)\textsuperscript{\cite{eckhardt2022}}.
SOAT1 knockdown suppresses lipogenesis and tumor growth in vivo\textsuperscript{\cite{liu2021b}}, and pharmacological inhibition of cholesterol esterification blocks prostate cancer metastasis in mouse models by impairing Wnt/$\beta$-catenin
signaling\textsuperscript{\cite{yue2018}}. The mechanistic coherence is further supported by the coupling of AR-driven programs to cholesterol and lipid metabolism, including evidence that metastatic prostate cancers can upregulate androgen synthesizing
enzymes enabling de novo androgen production from cholesterol even at castrate testosterone levels\textsuperscript{\cite{mohler2011}}. That PATH recovers this SOAT1-linked module from genomic alteration profiles alone, without gene expression data, supports
the model's capacity to capture biologically meaningful co variation embedded in multi omic inputs.

\FloatBarrier
\begin{figure*}[t!]
  \centering
  \includegraphics[width=\linewidth]{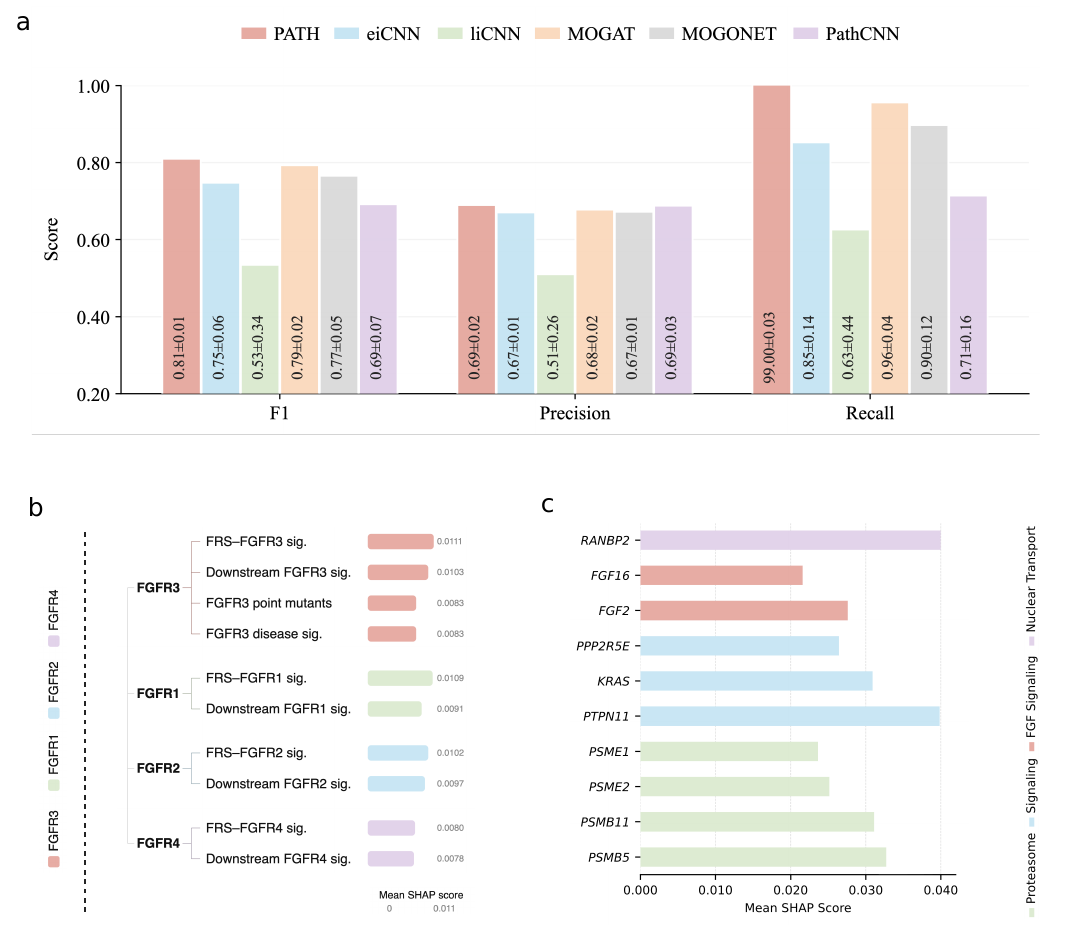}
  \caption{\textbf{PATH improves BLCA stage classification and highlights FGFR-related pathway and gene signals.}
  \textbf{(a)} Mean $\pm$ standard deviation of F1 score, precision, and recall across five-fold cross-validation for early versus late stage classification in BLCA. PATH achieves the best F1 (0.81$\pm$0.01) and perfect recall (1.00$\pm$0.00).
  \textbf{(b)} Pathway-level SHAP heatmap showing pairwise interactions among FGFR1--FGFR4 signaling axes and related downstream pathways. Higher scores indicate FGFR3-centered crosstalk associated with late-stage BLCA.
  \textbf{(c)} Gene-level SHAP scores for the top predictive genes grouped by pathway. FGF2, FGF16, RANBP2, and proteasome genes emerge as major contributors, indicating coordinated molecular programs captured by PATH.}
  \label{fig:blca_analysis}
\end{figure*}
\FloatBarrier

\subsection{PATH reveals pathway and gene-level features associated with early and late stage BLCA}
Before examining the biological signals learned by PATH, we first assessed its predictive performance for early versus late stage BLCA classification. As shown in Fig.~\ref{fig:blca_analysis}a, PATH achieved the strongest overall performance across five fold cross-validation, with the highest F1 score (0.81$\pm$0.02) and perfect recall (99.00$\pm$0.03). This pattern suggests that the model is particularly effective at identifying late stage tumors while maintaining competitive precision. Having established this predictive advantage, we next examined which pathways and genes contributed most strongly to the model's decisions.

\subsubsection*{BLCA Pathway Analysis: FGF/FGFR Signaling}

The PATH model identified ten top-ranked pathways in BLCA, all within the FGF/FGFR 
signaling family grouped by receptor subtype (FGFR1 to FGFR4). Six of these have 
strong literature support for a role in bladder cancer progression. The four FGFR3 
pathways are the most established. Activating FGFR3 mutations occur in approximately 
70\% of non-muscle invasive and approximately 15\% of muscle invasive bladder cancers, 
driving proliferation and survival through FRS2$\alpha$ mediated RAS/MAPK and PI3K/AKT 
signaling \cite{zheng2024fgfr3myc}. An FGFR3/MYC positive feedback loop further 
sustains tumor growth, and the phase III THOR trial confirmed clinical relevance, 
with erdafitinib extending overall survival over chemotherapy in FGFR3-altered 
advanced bladder cancer \cite{kwon2024fgfr}. The two FGFR1 pathways are also 
validated: FGFR1 activation induces epithelial to mesenchymal transition in bladder 
cancer cells via MAPK/PLCg/COX-2 signaling, promoting invasion and migration 
\cite{tomlinson2012fgfr1emt}, and FGFR1 expression is elevated in higher grade 
muscle invasive tumors where FGFR3 mutations are less frequent. The four remaining 
pathways, two FGFR2 and two FGFR4, require further investigation. FGFR2 behaves 
more as a tumor suppressor in bladder cancer, with low expression linked to worse 
prognosis and its re-expression reducing tumor growth preclinically 
\cite{knowles2012fgfr}. FGFR4 has no meaningful bladder-specific evidence, and 
FGFR4 genetic variants showed no significant association with tumor stage, grade, or 
metastasis in urothelial carcinoma \cite{huang2020fgfr4}. These signals likely 
reflect pancancer training data contributions and warrant dedicated BLCA functional 
studies.

\subsubsection*{PATH-Identified Gene Modules Capture Established and Emerging Drivers of Bladder Cancer Biology}

Among the top 10 PATH-identified genes, six have direct or indirect support in bladder cancer (BLCA) literature, while two require further investigation in the BLCA-specific context.

\paragraph{Directly supported in BLCA.}
FGF2 and FGF16 are ligands of the FGF receptor family, which is among the most frequently altered pathways in bladder cancer. FGF2 overexpression is enriched in invasive bladder tumors and correlates with poor prognosis, promoting tumor cell invasiveness and proliferation \cite{tsichlis2017}. FGF16 activates FGFR3, a receptor with well established roles in both non-muscle-invasive and muscle-invasive BLCA \cite{noeraparast2024}. PSMB5, a proteasome subunit, is the most directly validated proteasome gene in BLCA: knockdown of \textit{PSMB5} in BLCA cell lines inhibited proliferation and migration while promoting apoptosis, with machine learning applied to TCGA-BLCA data identifying it as a key risk gene linked to distinct molecular subtypes \cite{chen2025}.

\paragraph{Indirectly supported.}
KRAS and PTPN11 are well validated components of the RAS/MAPK pathway. In bladder cancer, RAS-MAPK is one of the two most important signaling axes in urothelial tumorigenesis, with RAS mutations detected across both early and late disease stages \cite{kompier2010}. PSME1 and PSME2 show pancancer prognostic relevance: hierarchical clustering of TCGA data identified \textit{PSME1} and \textit{PSME2} among a cluster with notably high expression in urologic cancers, with significant impact on overall survival when underexpressed \cite{larsson2022}.

The remaining two genes, PPP2R5E and RANBP2, have not been functionally characterized in BLCA to date. The detection of RANBP2–FGFR1 fusions in urothelial carcinoma provides a basis for future studies examining their role in bladder cancer progression.
\section{Discussion}

In this study, we developed PATH, a pathway-aware graph transformer that improves cancer progression prediction while also providing biologically interpretable outputs at the gene, pathway, and pathway to pathway levels. Across the pancancer, prostate cancer, and BLCA tasks, PATH consistently showed strong predictive performance relative to established pathway based and multi omics baselines (Figs.~\ref{fig:pancancer_performance}, \ref{fig:prostate_cancer}a,b, and \ref{fig:blca_analysis}a). These results suggest that modeling cancer progression through coordinated pathway organization, rather than treating pathways as isolated feature groups, can improve both predictive accuracy and interpretability.


A central finding of this work is that pathway crosstalk carries information that is not fully captured by static pathway summaries alone. Existing pathway informed models have improved interpretability by aggregating genes into biologically meaningful units, but they usually rely on fixed pathway features or curated pathway hierarchies. 
Our results suggest that this is not enough for studying progression related transitions such as metastasis, where the important signal may lie not only in which pathways are altered, but also in how pathways interact and reorganize across disease states \cite{valastyan2011tumor,dongre2019new,lambert2017emerging,creixell2015pathway}. By learning patient specific pathway representations from mutation and CNV inputs and allowing soft departures from the curated Reactome graph, PATH preserves biological prior knowledge while remaining flexible enough to detect context specific rewiring (Fig.~\ref{fig:overview}) \cite{gillespie2022reactome}.

The pancancer analysis provides the clearest example of this advantage. This was the most heterogeneous setting in the study, yet PATH achieved the strongest overall discrimination and learned latent representations that clearly separated primary from metastatic tumors (Fig.~\ref{fig:pancancer_performance}). More importantly, its pathway analyses converged on a coherent FGFR/MAPK centered signaling program associated with metastasis (Fig.~\ref{fig:pancancer_pathway_analysis}a,d). The prominence of Sprouty regulation of FGF signaling, negative regulation of FGFR pathways, and downstream MAPK modules suggests that metastatic progression is reflected not only by pathway activity, but also by coordinated reorganization of growth factor signaling \cite{hanahan2011,dent2003,hanafusa2002,helsten2016,davies2002,yao2019}. This interpretation is further supported by the crosstalk rewiring analysis, which showed metastasis associated gains in L1CAM related interactions and in FGFR2/Sprouty associated regulation (Fig.~\ref{fig:pancancer_pathway_analysis}c), consistent with literature linking L1CAM to invasion and metastasis and Sprouty mediated feedback loss to enhanced FGFR-ERK signaling \cite{kiefel2012,ganesh2020,xu2017sprouty,haglund2019}. Taken together, these findings suggest that PATH is capturing progression associated pathway organization rather than only isolated genomic alterations.

The model-inferred novel pathway interactions are also notable. In the pancancer setting, PATH recovered new links between extracellular matrix organization and the circadian clock, and between junction-related pathways and cilium assembly (Fig.~\ref{fig:pancancer_pathway_analysis}b). These connections are biologically plausible because extracellular matrix signaling can modulate circadian behavior, and cell polarity, junctional structure, and ciliary organization are closely linked in epithelial systems \cite{streuli2019,meng2023,tates2018,sfakianos2007,holst2011}. We interpret these results cautiously. They should not be treated as established biology, but they do show that the model can generate testable hypotheses beyond the edges already present in curated databases such as Reactome \cite{gillespie2022reactome}. This is an important practical strength of the framework.

The prostate cancer results show that PATH also remains informative in a disease focused setting with well established drivers. PATH again outperformed the comparison models (Fig.~\ref{fig:prostate_cancer}a,b), and the gene-level analysis identified \textit{AR} as the dominant metastatic signal (Fig.~\ref{fig:prostate_cancer}c), which is fully consistent with the central role of androgen receptor reactivation in advanced prostate cancer \cite{takeda2020,quigley2018}. The additional metastatic genes highlighted by the model, including \textit{RPTOR}, \textit{EIF4E}, \textit{AKT1}, \textit{PTEN}, and \textit{PTPN11}, converge on mTOR signaling, translational control, and the PI3K/AKT/PTEN axis, all of which are strongly linked to aggressive prostate cancer biology \cite{furic2010,dua2018,cotter2018,yoshimoto2013,zhang2015}. The gene module view further improves interpretability by grouping these signals into hormone signaling, GPI biosynthesis, and lipid metabolism programs (Fig.~\ref{fig:prostate_cancer}d). Among these, the SOAT1-associated lipid metabolism module is especially compelling because it fits known links among cholesterol handling, metastatic progression, and sustained AR related biology \cite{eckhardt2022,liu2021b,yue2018,mohler2011}. By contrast, the GPI biosynthesis module should be interpreted more cautiously. GPI-anchored proteins are broadly implicated in malignancy, but the specific contribution of these alterations to prostate cancer metastasis remains unclear \cite{lauber2020}.

The BLCA analysis extends the framework to stage classification and shows that PATH can recover meaningful signals in a different clinical context. PATH achieved the best overall classification performance and perfect recall for late stage disease in cross-validation (Fig.~\ref{fig:blca_analysis}a). Its pathway level outputs were concentrated in the FGF/FGFR family (Fig.~\ref{fig:blca_analysis}b), especially around FGFR3 centered signaling, which agrees with the known importance of FGFR3 alterations in bladder cancer biology and therapy \cite{zheng2024fgfr3myc,kwon2024fgfr}. The model also detected FGFR1 related signals, which is notable given prior evidence linking FGFR1 to invasive behavior and EMT-like changes in bladder cancer \cite{tomlinson2012fgfr1emt}. At the gene-level, PATH highlighted \textit{FGF2}, \textit{FGF16}, \textit{PSMB5}, \textit{KRAS}, and \textit{PTPN11} (Fig.~\ref{fig:blca_analysis}c), several of which have direct or indirect support in BLCA biology \cite{tsichlis2017,noeraparast2024,chen2025,kompier2010,peng2024ptpn11,larsson2022}. At the same time, the FGFR2 related, FGFR4 related, \textit{PPP2R5E} related, and \textit{RANBP2} related findings are less established in a bladder specific setting and should be viewed as potentially interesting leads rather than confirmed disease drivers \cite{knowles2012fgfr,huang2020fgfr4,saezayala2024ranbp2}.

This study has several limitations. First, the evaluation is based on retrospective datasets analyzed with cross-validation rather than external independent cohorts. Although the performance trends were consistent across all three tasks, broader validation will be needed to assess robustness across datasets, platforms, and clinical settings. Second, the biological conclusions are inference based. PATH identifies pathways, genes, and network changes that are statistically and biologically plausible, but these results do not establish causality and should be followed by targeted experimental validation. Third, the model currently integrates mutation and CNV only. This was sufficient to recover meaningful pathway programs, and the ablation analysis in prostate cancer suggests that combining both modalities is beneficial (Fig.~\ref{fig:prostate_cancer}e), but transcriptional, epigenetic, and microenvironmental factors are also likely to shape pathway activity and could further improve interpretation. Finally, PATH depends on existing pathway definitions and graph priors. Reactome provides a strong biological foundation, but curated pathway resources remain incomplete, especially for disease specific and context specific rewiring \cite{gillespie2022reactome,creixell2015pathway}.

Overall, our findings suggest that PATH provides a useful framework for pathway based genomic modeling of cancer progression. It preserves gene-level information, learns pathway representations in a patient specific way, and captures pathway crosstalk without discarding biological prior knowledge. This makes the model useful not only for classification, but also for generating mechanistic hypotheses about how signaling programs reorganize during disease progression. Future work should test PATH in external cohorts, extend it to additional omics layers, and experimentally examine the pathway interactions and gene modules highlighted here. More broadly, our results support the view that cancer progression is often better understood as a change in coordinated pathway organization than as the effect of single altered genes or isolated pathways.
\section{Methods}

\subsection{Data preprocessing and cohort description}

We evaluated PATH on three independent cancer cohorts spanning distinct classification tasks. The \textit{prostate cancer cohort} ($n = 1{,}011$; SU2C/PCF Dream Team, 333 primary and 678 metastatic samples)\cite{elmarakeby2021pnet} was used for primary versus metastatic classification. The \textit{pancancer cohort} spanning 33 TCGA tumour types (TCGA PANCAN, accessed via UCSC Xena)\cite{goldman2020xena,tcga2013comprehensive} was used for primary versus metastatic classification across cancer types. The \textit{bladder cancer cohort} (TCGA BLCA)\cite{goldman2020xena} was used for early stage (stage~I to II) versus late stage (stage~III to IV) classification, testing PATH's ability to distinguish tumour progression states beyond the primary to metastatic dichotomy. For each cohort, somatic mutation calls were encoded as binary indicators ($m_{ig} \in \{0,1\}$) and copy number variation (CNV) values were retained as continuous scores; the two modalities were merged additively per gene per patient. Genes with alteration frequency below 1\% were removed, and a globally consistent gene vocabulary of $G$ genes was constructed to support gene specific representation learning.

CNV values were standardised to zero mean and unit variance using z-score normalisation. To prevent information leakage, the normalisation statistics (mean $\mu$ and standard deviation $\sigma$) were computed exclusively from the training partition of each cross-validation fold, and the same statistics were applied to transform the validation and test partitions:
\begin{equation}
\tilde{c}_{ig} = \frac{c_{ig} - \mu_g^{(\text{train})}}{\sigma_g^{(\text{train})} + \epsilon}
\label{eq:zscore}
\end{equation}
where $\epsilon = 10^{-8}$ prevents division by zero. Binary mutation indicators were used without further normalisation.

Pathway definitions and gene membership were obtained from the Reactome database (2,484 pathways; 11,375 unique genes)\cite{elmarakeby2021pnet}, filtered to pathways containing $\geq 15$ genes present in the processed profiles. Only pathways with at least one gene mapped to the mutation/CNV gene vocabulary were retained, yielding $P$ pathways. Each pathway $p$ is associated with a gene membership set $\mathcal{G}_p \subseteq \{1, \ldots, G\}$ of variable size $|\mathcal{G}_p|$. Pathway to pathway interaction structure was derived from shared gene membership using Jaccard similarity, yielding a sparse adjacency matrix $\mathbf{A} \in \mathbb{R}^{P \times P}$ used as the biological graph prior. The adjacency matrix was symmetrised and normalised:
\begin{equation}
\mathbf{A} \leftarrow \frac{\tfrac{1}{2}(\mathbf{A} + \mathbf{A}^\top)}{\max(\mathbf{A}) + \epsilon}
\label{eq:adj_norm}
\end{equation}
In the full-graph configuration, $\mathbf{A}$ is replaced by a complete graph with all off-diagonal entries set to 1 and self-loops set to 0, allowing unrestricted pathway to pathway communication. Full details of cohort characteristics, omics filtering, gene vocabulary construction, and pathway network assembly are provided in Supplementary Section~\ref{sec:supp_data_preprocessing}.

\subsection{Model architecture}

PATH is a four stage deep learning architecture that transforms raw per gene molecular measurements into a patient level classification through successive levels of biological abstraction (Fig.~\ref{fig:overview}): (i) gene-level representation learning via Feature wise Linear Modulation (FiLM), (ii) within pathway attention pooling to construct pathway tokens, (iii) pathway level message passing through an edge aware graph transformer with soft structural masking, and (iv) attention weighted readout with classification.

\subsubsection*{Stage 1: Gene-level representation learning via FiLM}

Each gene $g \in \{1, \ldots, G\}$ is assigned a learnable embedding vector $\mathbf{e}_g \in \mathbb{R}^d$, where $d$ is the model dimension (default $d = 64$). Embeddings are initialised from a normal distribution with standard deviation 0.02. Rather than concatenating omic features directly, which would conflate gene identity with patient specific measurements, PATH employs Feature wise Linear Modulation (FiLM)\cite{perez2018film} to condition each gene embedding on the patient's molecular observations at that locus.

For patient $i$ and gene $g$, the binary mutation indicator $m_{ig}$ and normalised CNV value $\tilde{c}_{ig}$ are stacked into a 2-dimensional input vector $\mathbf{x}_{ig} = [m_{ig},\, \tilde{c}_{ig}]^\top$. A two-layer MLP with GELU activation and hidden dimension $d_h = 64$ maps this input to $2d$-dimensional modulation parameters:
\begin{equation}
[\boldsymbol{\gamma}_{ig}^{(\text{raw})},\, \boldsymbol{\beta}_{ig}] = \text{MLP}_{\text{FiLM}}(\mathbf{x}_{ig}) = \mathbf{W}_2\, \text{GELU}(\mathbf{W}_1 \mathbf{x}_{ig} + \mathbf{b}_1) + \mathbf{b}_2
\label{eq:film_mlp}
\end{equation}
where $\mathbf{W}_1 \in \mathbb{R}^{d_h \times 2}$, $\mathbf{W}_2 \in \mathbb{R}^{2d \times d_h}$, and the output is split into scale and shift components $\boldsymbol{\gamma}_{ig}^{(\text{raw})}, \boldsymbol{\beta}_{ig} \in \mathbb{R}^d$. The scale parameter is constrained to be strictly positive via softplus activation:
\begin{equation}
\boldsymbol{\gamma}_{ig} = \text{softplus}(\boldsymbol{\gamma}_{ig}^{(\text{raw})}) + 10^{-3}
\label{eq:film_gamma}
\end{equation}
The gene representation is then computed as:
\begin{equation}
\mathbf{h}_{ig} = \boldsymbol{\gamma}_{ig} \odot \mathbf{e}_g + \boldsymbol{\beta}_{ig}
\label{eq:film}
\end{equation}
where $\odot$ denotes element-wise (Hadamard) multiplication. This modulation scheme allows the model to learn how somatic mutations and copy number alterations jointly reshape gene identity in a patient specific manner: $\boldsymbol{\gamma}_{ig}$ scales the magnitude of each embedding dimension while $\boldsymbol{\beta}_{ig}$ shifts the representation in directions determined by the patient's molecular profile.

To ensure stable training, the FiLM MLP is initialised to produce near-identity modulation at the start of training. Specifically, the final layer weights $\mathbf{W}_2$ are initialised to zero and the biases are set such that $\boldsymbol{\gamma}^{(\text{raw})} \approx 0$ (yielding $\boldsymbol{\gamma} \approx \text{softplus}(0) + 10^{-3} \approx 0.694$) and $\boldsymbol{\beta} \approx \mathbf{0}$, so that initial gene representations closely approximate the raw embeddings.

\subsubsection*{Stage 2: Pathway token construction via attention pooling}

The gene representations belonging to each pathway must be aggregated into a single pathway level token. Because pathway gene sets vary in size (from $|\mathcal{G}_p| = 15$ to several hundred) and not all member genes contribute equally to metastatic phenotypes, PATH employs a learnable attention based pooling mechanism rather than simple averaging.

For each pathway $p$ containing gene indices $\mathcal{G}_p$, the corresponding gene representations $\{\mathbf{h}_{ig}\}_{g \in \mathcal{G}_p}$ are gathered from the full gene representation matrix. Attention scores are computed via a Bahdanau-style\cite{bahdanau2015attention} mechanism with a pathway-specific bias:
\begin{equation}
s_{ipg} = \mathbf{u}^\top \tanh(\mathbf{W} \mathbf{h}_{ig} + \mathbf{b}_p), \quad g \in \mathcal{G}_p
\label{eq:attention_score}
\end{equation}
where $\mathbf{W} \in \mathbb{R}^{d \times d}$ is a shared linear projection (with bias), $\mathbf{u} \in \mathbb{R}^d$ is a learnable context vector (initialised with standard deviation 0.02), and $\mathbf{b}_p \in \mathbb{R}^d$ is a pathway-specific bias vector (initialised to zero) that allows the pooling mechanism to adapt to the functional context of each pathway. Scores for genes not belonging to pathway $p$ are masked to $-\infty$ before softmax normalisation, ensuring that attention is restricted to member genes:
\begin{equation}
\alpha_{ipg} = \frac{\exp(s_{ipg})}{\sum_{g' \in \mathcal{G}_p} \exp(s_{ipg'})}, \quad g \in \mathcal{G}_p
\label{eq:attention_weights}
\end{equation}
The pathway token is computed as the attention-weighted sum:
\begin{equation}
\mathbf{z}_{ip} = \sum_{g \in \mathcal{G}_p} \alpha_{ipg}\, \mathbf{h}_{ig}
\label{eq:pathway_token}
\end{equation}
This produces a matrix $\mathbf{Z}_i \in \mathbb{R}^{P \times d}$ of pathway tokens for patient $i$, each summarising the patient's molecular state at the pathway level. The attention weights $\alpha_{ipg}$ are retained as interpretability handles, revealing which genes drive the representation of each pathway for a given patient.

\subsubsection*{Stage 3: Pathway-aware graph transformer}

\paragraph{Laplacian positional encoding.}
To endow the graph transformer with awareness of each pathway's structural position within the pathway interaction network, PATH computes Laplacian positional encodings (LPE)\cite{dwivedi2020benchmarking}. The symmetric normalised graph Laplacian is defined as:
\begin{equation}
\mathbf{L} = \mathbf{I} - \mathbf{D}^{-1/2} \mathbf{A} \mathbf{D}^{-1/2}
\label{eq:laplacian}
\end{equation}
where $\mathbf{D} = \text{diag}(d_1, \ldots, d_P)$ is the degree matrix with $d_p = \sum_q A_{pq}$ (clamped to a minimum of $10^{-8}$ for numerical stability). An eigendecomposition $\mathbf{L} = \mathbf{V} \boldsymbol{\Lambda} \mathbf{V}^\top$ yields eigenvectors whose first $k$ columns (default $k = 16$) serve as positional features. These are concatenated with the normalised node degree and linearly projected to the model dimension:
\begin{equation}
\text{PE}_p = \mathbf{W}_{\text{PE}} \left[\mathbf{v}_p^{(1)}, \ldots, \mathbf{v}_p^{(k)},\, \frac{d_p}{P + \epsilon}\right] + \mathbf{b}_{\text{PE}}
\label{eq:pe}
\end{equation}
where $\mathbf{W}_{\text{PE}} \in \mathbb{R}^{d \times (k+1)}$ is a learnable projection. During training, random sign-flip augmentation is applied independently to each eigenvector dimension to encourage invariance to the inherent sign ambiguity of eigenvectors:
\begin{equation}
\tilde{\mathbf{v}}_p^{(j)} = r_j \cdot \mathbf{v}_p^{(j)}, \quad r_j \sim \text{Uniform}\{-1, +1\}, \quad j = 1, \ldots, k
\label{eq:sign_flip}
\end{equation}
The positional encoding is added to each pathway token to form the input to the graph transformer:
\begin{equation}
\mathbf{x}_{ip}^{(0)} = \mathbf{z}_{ip} + \text{PE}_p
\label{eq:pe_addition}
\end{equation}

\paragraph{Edge-aware graph transformer blocks.}
The augmented pathway tokens $\{\mathbf{x}_{ip}^{(0)}\}_{p=1}^{P}$ are processed through a stack of $L$ (default $L = 2$) edge aware graph transformer blocks inspired by Dwivedi and Bresson\cite{dwivedi2021generalization}. Unlike standard transformers, these blocks incorporate the pathway adjacency structure through two complementary mechanisms: soft structural masking and edge conditioned attention bias. The blocks also maintain and update edge features across layers.

\textit{Soft structural masking.} Rather than hard masking nonedge edges with $-\infty$ (which blocks gradient flow entirely), a soft penalty is applied to pathway pairs not connected in the Reactome-derived adjacency:
\begin{equation}
m_{pq}^{(\text{struct})} = \begin{cases}
0, & \text{if } A_{pq} > 0 \text{ or } p = q \\
-10.0, & \text{otherwise}
\end{cases}
\label{eq:soft_masking}
\end{equation}
This retains a strong inductive bias from the biological prior while allowing the model to discover novel pathway to pathway interactions through small but nonzero attention weights on masked edges. In the full-graph configuration, $m_{pq}^{(\text{struct})} = 0$ for all pairs.

\textit{Edge-conditioned attention bias.} The scalar adjacency weight $A_{pq}$ between each pathway pair is treated as a one-dimensional edge feature $\mathbf{e}_{pq} = [A_{pq}]$. At each transformer layer $\ell$, a linear projection maps the edge feature to per-head gain values:
\begin{equation}
\varphi_{pq}^{(h)} = \text{softplus}\!\left(w_h^{(\ell)} \cdot e_{pq}^{(\ell)} + b_h^{(\ell)}\right), \quad h = 1, \ldots, H
\label{eq:edge_gain}
\end{equation}
where $w_h^{(\ell)}$ and $b_h^{(\ell)}$ are learnable per-head parameters and $H = 4$ is the number of attention heads. The gains are aggregated into a single additive bias by taking the mean of the log-transformed per-head values:
\begin{equation}
\phi_{pq}^{(\ell)} = \frac{1}{H}\sum_{h=1}^{H} \log\!\left(\varphi_{pq}^{(h)} + \epsilon\right)
\label{eq:edge_bias}
\end{equation}
The final attention mask combines the structural mask with the edge bias:
\begin{equation}
M_{pq}^{(\ell)} = m_{pq}^{(\text{struct})} + \phi_{pq}^{(\ell)}
\label{eq:attention_mask}
\end{equation}
This mask is broadcast across all $H$ heads and added to the scaled dot product attention logits, so that multi head self attention at layer $\ell$ is computed as:
\begin{equation}
\text{MHSA}^{(\ell)}(\mathbf{X}) = \text{softmax}\!\left(\frac{\mathbf{Q}^{(\ell)} {\mathbf{K}^{(\ell)}}^\top}{\sqrt{d/H}} + \mathbf{M}^{(\ell)}\right) \mathbf{V}^{(\ell)}
\label{eq:multihead_attention}
\end{equation}
where $\mathbf{Q}^{(\ell)}, \mathbf{K}^{(\ell)}, \mathbf{V}^{(\ell)}$ are the query, key, and value projections of the input $\mathbf{X}^{(\ell-1)}$ at layer $\ell$.

\textit{Residual connections and normalisation.} Each transformer block applies post attention and post FFN residual connections with batch normalisation\cite{ioffe2015batch}, applied per token by reshaping the tensor from $(B, P, d)$ to $(B \cdot P, d)$ before normalisation. We found that batch normalisation provided more stable training than layer normalisation at the batch sizes used. A position wise feed forward network (FFN) with expansion factor 4 and GELU activation\cite{hendrycks2016gelu} is applied after attention:
\begin{align}
\hat{\mathbf{X}}^{(\ell)} &= \text{BN}\!\left(\mathbf{X}^{(\ell-1)} + \text{MHSA}^{(\ell)}(\mathbf{X}^{(\ell-1)})\right) \label{eq:residual_attn} \\
\mathbf{X}^{(\ell)} &= \text{BN}\!\left(\hat{\mathbf{X}}^{(\ell)} + \text{FFN}^{(\ell)}(\hat{\mathbf{X}}^{(\ell)})\right) \label{eq:residual_ffn}
\end{align}
where $\text{FFN}^{(\ell)}(\mathbf{x}) = \mathbf{W}_2^{(\ell)}\, \text{Dropout}\!\big(\text{GELU}(\mathbf{W}_1^{(\ell)} \mathbf{x} + \mathbf{b}_1^{(\ell)})\big) + \mathbf{b}_2^{(\ell)}$, with $\mathbf{W}_1^{(\ell)} \in \mathbb{R}^{4d \times d}$ and $\mathbf{W}_2^{(\ell)} \in \mathbb{R}^{d \times 4d}$. Dropout with rate $p_{\text{drop}} = 0.2$ is applied after both the GELU activation and the second linear layer.

\textit{Edge feature update.} Edge features are simultaneously updated at each layer through a parallel two-layer MLP with GELU activation and batch normalisation:
\begin{equation}
\mathbf{e}_{pq}^{(\ell+1)} = \text{BN}\!\left(\text{MLP}_{\text{edge}}^{(\ell)}(\mathbf{e}_{pq}^{(\ell)})\right)
\label{eq:edge_update}
\end{equation}
This allows the model to refine its representation of pathway to pathway relationships across layers, enabling dynamic edge reweighting during message passing.

\subsubsection*{Stage 4: Attention-weighted readout and classification}

After the final transformer layer, a global patient representation is obtained via attention weighted pooling over pathway tokens. A two layer attention network with $\tanh$ activation computes pathway importance scores:
\begin{equation}
w_{ip} = \text{softmax}_p\!\left(\mathbf{v}^\top \tanh\!\left(\mathbf{W}_{\text{attn}} \mathbf{x}_{ip}^{(L)}\right)\right)
\label{eq:pathway_weights}
\end{equation}
where $\mathbf{W}_{\text{attn}} \in \mathbb{R}^{(d/2) \times d}$, $\mathbf{v} \in \mathbb{R}^{d/2}$, and the softmax is taken over all $P$ pathways. The patient level representation is the weighted sum:
\begin{equation}
\mathbf{g}_i = \sum_{p=1}^{P} w_{ip}\, \mathbf{x}_{ip}^{(L)}
\label{eq:global_pooling}
\end{equation}
This representation is passed through a classification head consisting of a linear projection ($d \to d$), batch normalisation, GELU activation, dropout ($p_{\text{drop}} = 0.2$), and a final linear projection to two class logits:
\begin{equation}
\hat{\mathbf{y}}_i = \mathbf{W}_{\text{out}}\, \text{Dropout}\!\big(\text{GELU}(\text{BN}(\mathbf{W}_{\text{cls}} \mathbf{g}_i + \mathbf{b}_{\text{cls}}))\big) + \mathbf{b}_{\text{out}}
\label{eq:classification}
\end{equation}
where $\hat{\mathbf{y}}_i \in \mathbb{R}^2$ are the logits for the primary and metastatic classes. The predicted probability of metastatic status is obtained by applying softmax to the logits. The pathway attention weights $w_{ip}$ directly quantify each pathway's contribution to the classification decision and are used in downstream interpretability analyses (see Section~\ref{sec:interpretability}).

A singleton-safe batch normalisation variant is employed throughout the model to handle edge cases where a mini-batch may contain a single sample (e.g., the final mini-batch of an epoch). When the batch size is 1, running statistics are used in place of batch statistics to avoid undefined variance estimates.

\subsection{Training and evaluation}

\subsubsection*{Cross-validation strategy}

Model performance was assessed using 5 fold stratified cross-validation repeated under two independent random seeds (42 and 123), yielding a total of 10 train/test runs per cohort. For each fold, data were partitioned using \texttt{StratifiedKFold} with 20\% of patients held out as the test set and the remaining 80\% further split into training (90\%) and validation (10\%) partitions using stratified sampling, yielding approximate per fold proportions of 72\% training, 8\% validation, and 20\% test. Class distributions were maintained in all partitions. The two random seeds produce different fold assignments and model initialisations, providing a more robust estimate of performance variability than a single 5 fold run. Within each seed, per fold random seeds were set deterministically (seed $= s + k$ for base seed $s \in \{42, 123\}$ and fold $k \in \{1,\ldots,5\}$), controlling all sources of randomness including weight initialisation, data shuffling, and dropout masks. Fold assignments were serialised to disk and reloaded across experiments to ensure reproducibility. All metrics are reported as mean $\pm$ standard deviation across all 10 test folds unless otherwise noted.

\subsubsection*{Loss function}

The model was trained with cross-entropy loss augmented by inverse-frequency class weights to address class imbalance between primary and metastatic samples:
\begin{equation}
\mathcal{L} = -\frac{1}{N}\sum_{i=1}^{N} w_{y_i} \left[ y_i \log \hat{p}_i + (1 - y_i) \log(1 - \hat{p}_i) \right]
\label{eq:loss}
\end{equation}
where $\hat{p}_i = \text{softmax}(\hat{\mathbf{y}}_i)_1$ is the predicted metastatic probability and $w_{y_i}$ is the class weight for sample $i$, computed via scikit-learn's \texttt{compute\_class\_weight} with the \texttt{balanced} strategy:
\begin{equation}
w_c = \frac{N}{C \cdot n_c}, \quad c \in \{0, 1\}
\label{eq:class_weight}
\end{equation}
where $N$ is the total number of training samples, $C = 2$ is the number of classes, and $n_c$ is the count of class $c$. Optionally, focal loss\cite{lin2017focal} can be applied to further down-weight well-classified examples:
\begin{equation}
\mathcal{L}_{\text{focal}} = -\frac{1}{N}\sum_{i=1}^{N} w_{y_i}\, (1 - p_{t,i})^{\gamma}\, \text{CE}(y_i, \hat{p}_i)
\label{eq:focal_loss}
\end{equation}
where $p_{t,i}$ is the predicted probability assigned to the true class and $\gamma = 2.0$ is the focusing parameter.

\subsubsection*{Optimisation}

Parameters were optimised with AdamW\cite{loshchilov2019adamw} (learning rate $\eta = 10^{-4}$, weight decay $\lambda = 5 \times 10^{-4}$). Training continued for a maximum of 200 epochs with batch size 16. Gradient norms were clipped to a maximum of 2.0 to prevent training instabilities. Early stopping with patience of 25 epochs monitored validation AUROC, with a minimum of 50 training epochs enforced to allow sufficient initial learning before early stopping could trigger. The model checkpoint achieving the highest validation AUROC was retained for test-set evaluation.

\subsubsection*{Decision threshold calibration}

At evaluation time, predicted probabilities were converted to binary predictions using a threshold optimised on the validation set. Specifically, we computed the precision and recall curve on validation predictions and selected the threshold $\tau^*$ maximising the F1 score:
\begin{equation}
\tau^* = \arg\max_\tau \frac{2 \cdot \text{Precision}(\tau) \cdot \text{Recall}(\tau)}{\text{Precision}(\tau) + \text{Recall}(\tau) + 10^{-12}}
\label{eq:threshold}
\end{equation}
This calibrated threshold was then applied to the test set, providing more accurate binary predictions than the default threshold of 0.5, particularly under class imbalance.

\subsubsection*{Performance metrics}

Classification performance was evaluated using six complementary metrics: area under the receiver operating characteristic curve (AUROC), area under the precision and recall curve (AUPRC), binary F1 score, precision, recall, and overall accuracy. AUROC was used as the primary metric for early stopping and model selection. AUPRC was reported alongside AUROC because it is more informative under class imbalance, as it focuses on the positive class. All metrics were computed per fold and reported as mean $\pm$ standard deviation across all 10 test folds (5 folds $\times$ 2 random seeds). Aggregate ROC curves and precision and recall curves were generated by interpolating per fold curves onto a common grid of 100 equally spaced points, with $\pm 1$ standard deviation bands visualised to assess cross fold and cross seed consistency. Aggregate confusion matrices were computed as the element wise mean and standard deviation across all 10 folds.

\subsubsection*{Contribution of mutation and copy number signals}

To quantify the independent contribution of each omics modality, we performed an ablation analysis using the same 5 fold stratified cross-validation framework and identical fold assignments as the primary model. Three configurations were evaluated: the full model using both mutation and CNV inputs, a mutation only variant (CNV inputs set to zero), and a CNV only variant (mutation inputs set to zero). Setting the non evaluated modality to zero rather than removing it preserves the model architecture, parameter count, and training protocol, thereby isolating the effect of input modality without introducing confounding changes in model capacity. Performance differences across configurations were assessed using the same six metrics described above.

\subsection{Interpretability analysis}
\label{sec:interpretability}

PATH provides three complementary levels of interpretability: pathway level importance ranking, gene-level attention attribution, and pathway to pathway crosstalk analysis. All interpretability outputs are computed from the trained model in inference mode (dropout disabled, batch normalisation using running statistics).

\subsubsection*{Pathway-level SHAP attribution}

Pathway-level SHAP attribution was computed after model training in each cross-validation fold using SHAP GradientExplainer applied to the class-1 logit. SHAP values were obtained separately for mutation and copy-number variation inputs and summed at the gene level to generate a unified signed attribution for gene $g$ in sample $i$,
\[
\phi_{ig} = \phi^{\mathrm{mut}}_{ig} + \phi^{\mathrm{cnv}}_{ig}.
\]
For each pathway $p$, containing gene set $G_p$, pathway attribution was defined as the mean signed SHAP value across all member genes,
\[
\Phi_{ip} = \frac{1}{|G_p|}\sum_{g \in G_p}\phi_{ig}.
\]

To quantify pathway shift toward the positive class, we calculated a differential pathway SHAP score as the difference between the mean signed pathway attribution in class 1 and class 0,
\[
\Delta_p = \mathbb{E}[\Phi_{ip} \mid y_i = 1] - \mathbb{E}[\Phi_{ip} \mid y_i = 0].
\]
Pathways were ranked by $\Delta_p$, such that larger positive values indicated stronger positive-class-associated pathway contribution. Ranking robustness across folds and random seeds was assessed by aggregating pathway recurrence, mean rank, and mean $\Delta_p$ across runs.

\subsubsection*{Identification of new pathway edges}

New pathway edges were identified by comparing the learned pathway--pathway interaction matrix with the prior adjacency matrix used to define the input graph. Directed edges among the top 20 rewired pathways were ranked by learned metastatic interaction weight. For each edge from pathway $i$ to pathway $j$, the learned weight $L_{ij}$ was taken from the aggregated metastatic interaction matrix and the prior weight $B_{ij}$ from the original adjacency matrix. Edges were classified as new when absent from the prior graph,
\[
\mathrm{new}_{ij} =
\begin{cases}
1, & B_{ij} \leq 0 \\
0, & B_{ij} > 0
\end{cases}
\]
The final edge table also included class-specific mean interaction strengths, differential interaction values, and false-discovery-rate-adjusted significance statistics from the edge-level comparison between metastatic and primary samples.

\subsubsection*{Differential pathway crosstalk rewiring}

Differential pathway crosstalk rewiring was inferred from class-specific pathway--pathway attention matrices learned by the graph transformer. For each directed pathway pair $(i,j)$, rewiring was defined as the difference between the aggregate metastatic and primary interaction strengths,
\[
\Delta_{ij} = \bar{A}^{\mathrm{met}}_{ij} - \bar{A}^{\mathrm{pri}}_{ij},
\]
where $\bar{A}^{\mathrm{met}}_{ij}$ and $\bar{A}^{\mathrm{pri}}_{ij}$ denote the mean metastatic and primary attention weights, respectively. We retained edges with the largest positive values of $\Delta_{ij}$ to define a focused rewiring subnetwork. To assess statistical support, each directed edge was matched to its edge-wise false-discovery-rate-adjusted $q$ value obtained from the primary-versus-metastatic comparison of sample-level interaction weights. Edges were considered significantly rewired at $q < 0.05$.

\subsubsection*{Gene-level SHAP attribution}

Gene-level SHAP attribution was computed after model training in each cross-validation fold using SHAP GradientExplainer applied to the class-1 logit. SHAP values were obtained separately for mutation and copy-number variation inputs and summed to generate a unified signed attribution for gene $g$ in sample $i$,
\[
\phi_{ig} = \phi^{\mathrm{mut}}_{ig} + \phi^{\mathrm{cnv}}_{ig}.
\]
To quantify class-associated gene contribution, we calculated a differential SHAP score for each gene as
\[
\Delta_g = \mathbb{E}[\phi_{ig} \mid y_i = 1] - \mathbb{E}[\phi_{ig} \mid y_i = 0].
\]
Genes were ranked by $\Delta_g$, with larger positive values indicating stronger positive-class-associated contribution. Ranking robustness across folds and random seeds was assessed by aggregating gene recurrence, mean rank, and mean $\Delta_g$ across runs.

\subsubsection*{Hierarchical hub organization analysis}

Hierarchical hub organization was derived from aggregated pathway-level SHAP signals and the prior pathway adjacency graph. First, for each pathway $p$, we computed an aggregate differential SHAP score across folds,
\[
\Delta_p = \mathbb{E}[\Phi_{ip}\mid y_i=1] - \mathbb{E}[\Phi_{ip}\mid y_i=0],
\]
and retained the positive component,
\[
s_p = \max(\Delta_p, 0).
\]
Using the pathway adjacency matrix $A$, we then defined a SHAP-informed crosstalk score between pathways $i$ and $j$ as
\[
E_{ij} = A_{ij}\, s_i s_j,
\]
such that an edge was emphasized only when the pathway pair was connected in the prior graph and both pathways showed positive class-associated SHAP signal.

For each pathway, a hub score was calculated as the sum of its outgoing crosstalk scores,
\[
H_i = \sum_j E_{ij}.
\]
Pathways were ranked by $H_i$ to identify the dominant SHAP-supported hub pathways, and for each hub, neighboring pathways were ordered by edge score to define a hierarchical organization of hub-to-neighbor relationships. These summarizes pathways that are not only individually shifted toward the positive class, but also occupy central positions within a connected network of similarly shifted pathways. This representation is meaningful because it prioritizes coordinated pathway programs supported jointly by attribution strength and known pathway topology, rather than isolated pathway effects alone.

\subsubsection*{Gene signature extraction}

For each pathway, signature genes were identified using the within pathway attention weights $\alpha_{ipg}$ (Eq.~\ref{eq:attention_weights}) from the pathway token construction stage. Class specific gene importance was computed by averaging attention weights across all samples of each class:
\begin{equation}
\bar{\alpha}_{pg}^{(c)} = \frac{1}{|\mathcal{S}_c|}\sum_{i \in \mathcal{S}_c} \alpha_{ipg}, \quad c \in \{\text{primary}, \text{metastatic}\}
\label{eq:gene_importance}
\end{equation}
where $\mathcal{S}_c$ is the set of patients belonging to class $c$. Differential gene importance ($\bar{\alpha}_{pg}^{(\text{met})} - \bar{\alpha}_{pg}^{(\text{pri})}$) identifies genes whose contributions to pathway activity differ between disease states.

We additionally employed Integrated Gradients (IG)\cite{sundararajan2017axiomatic} to attribute predictions to individual gene alterations. IG computes the path integral of gradients from a baseline input $\mathbf{x}^{(0)}$ (all zeros, representing a patient with no observed alterations) to the observed input $\mathbf{x}$:
\begin{equation}
\text{IG}_g(\mathbf{x}) = (x_g - x_g^{(0)}) \times \int_0^1 \frac{\partial f(\mathbf{x}^{(0)} + t(\mathbf{x} - \mathbf{x}^{(0)}))}{\partial x_g}\, dt
\label{eq:integrated_gradients}
\end{equation}
where $f$ is the model's metastatic-class logit. The integral was approximated using 50 Riemann steps. Class-specific IG scores were computed by averaging across samples within each class, and genes were ranked by the metastatic-minus-primary IG difference to identify genomic alterations most predictive of metastatic progression.

\subsection{Baseline methods}

We compared PATH against multiple baseline approaches:

\begin{itemize}
\item \textbf{PathCNN}\cite{oh2021pathcnn}: Pathway features via PCA, CNN-based classification.
\item \textbf{MOGAT} and \textbf{MOGONET}\cite{wang2021mogonet}: Multi-omics graph attention networks.
\item \textbf{eiCNN} and \textbf{liCNN}: CNN-based multi-omics integration.
\end{itemize}

All baselines were trained using the same 5 fold cross-validation framework with identical data splits, fold assignments, and random seeds (42 and 123) for fair comparison. To ensure that performance differences reflect architectural choices rather than hyperparameter tuning advantages, all baseline methods were trained using the same core hyperparameters as PATH (learning rate $10^{-4}$, weight decay $5 \times 10^{-4}$, batch size 16, dropout rate 0.2, and early stopping with patience of 25 epochs based on validation AUROC) unless a method's architecture required method specific adjustments (e.g., number of convolutional filters). Performance was evaluated using identical metrics across all methods.

\subsection{Software and reproducibility}

PATH was implemented in PyTorch (v2.0+). Data preprocessing, statistical analyses, and visualisations were performed using NumPy (v1.24+), pandas (v2.0+), scikit-learn (v1.3+), Matplotlib (v3.7+), and Seaborn (v0.12+). All random seeds were set deterministically (base seeds 42 and 123 for fold generation; derived per fold seeds for model training) controlling Python, NumPy, and PyTorch random number generators. Fold split indices for both random seeds were serialised and versioned. Trained model checkpoints, per sample predictions, and pathway importance rankings were saved for all 10 runs to support post hoc analysis. 

\subsection{Model complexity and computational considerations}

The total number of trainable parameters in PATH is:
\begin{equation}
|\Theta| = \underbrace{G \cdot d}_{\text{gene embeddings}} + \underbrace{2(2 d_h + d_h \cdot 2d)}_{\text{FiLM MLP}} + \underbrace{d^2 + P \cdot d + d}_{\text{pathway pool}} + \underbrace{L(12d^2 + 8d + H + 3)}_{\text{transformer}} + \underbrace{d^2/2 + 3d/2 + d^2 + d + 2d + 2}_{\text{readout + head}}
\label{eq:params}
\end{equation}
For the default configuration ($G \approx 5{,}000$, $P \approx 300$, $d = 64$, $d_h = 64$, $L = 2$, $H = 4$), this yields approximately 400K to 500K parameters, significantly fewer than dense multi omics models operating on the full gene space. The pathway based tokenisation reduces the sequence length from $G$ (thousands of genes) to $P$ (hundreds of pathways), making the $\mathcal{O}(P^2)$ self attention computation tractable without approximation.






 





\bibliographystyle{naturemag}
\bibliography{references}      



\section*{Acknowledgements (not compulsory)}

Acknowledgements should be brief, and should not include thanks to anonymous referees and editors, or effusive comments. Grant or contribution numbers may be acknowledged.

\section*{Author contributions statement}

Must include all authors, identified by initials, for example:
A.A. conceived the experiment(s),  A.A. and B.A. conducted the experiment(s), C.A. and D.A. analysed the results.  All authors reviewed the manuscript. 




\end{document}